\documentclass[10pt,twocolumn,letterpaper]{article}

\usepackage{iccv}
\usepackage{times}
\usepackage[dvipdfmx]{graphicx}
\usepackage{mediabb}
\usepackage{amsmath}
\usepackage{amssymb}

\usepackage{multirow}
\usepackage{subfigure}
\usepackage{algorithm}
\usepackage{algorithmic}

\newcommand{\argmin}{\mathop{\rm arg~min}\limits}

\usepackage[breaklinks=true,bookmarks=false]{hyperref}

\iccvfinalcopy 

\def\iccvPaperID{ 36} 

\begin{document}

\title{Binary-decomposed DCNN for accelerating computation and compressing model without retraining}

\author{Ryuji Kamiya\\
Chubu University\\
\and
Takayoshi Yamashita\\
Chubu University\\
\and
Mitsuru Ambai\\
Denso IT Laboratory\\
\and
Ikuro Sato\\
Denso IT Laboratory\\
\and
Yuji Yamauchi\\
Chubu University\\
\and
Hironobu Fujiyoshi\\
Chubu University
}

\maketitle

\begin{abstract}
Recent trends show recognition accuracy increasing even more profoundly.
Inference process of Deep Convolutional Neural Networks (DCNN) has a large number of parameters, requires a large amount of computation, and can be very slow.
The large number of parameters also require large amounts of memory.
This is resulting in increasingly long computation times and large model sizes.
To implement mobile and other low performance devices incorporating DCNN, model sizes must be compressed and computation must be accelerated.
To that end, this paper proposes Binary-decomposed DCNN, which resolves these issues without the need for retraining.
Our method replaces real-valued inner-product computations with binary inner-product computations in existing network models to accelerate computation of inference and decrease model size without the need for retraining.
Binary computations can be done at high speed using logical operators such as XOR and AND, together with bit counting.
In tests using AlexNet with the ImageNet classification task, speed increased by a factor of 1.79, models were compressed by approximately 80\%, and increase in error rate was limited to 1.20\%.
With VGG-16, speed increased by a factor of 2.07, model sizes decreased by 81\%, and error increased by only 2.16\%.
\end{abstract}

\section{Introduction}
Deep Convolutional Neural Networks (DCNN) realize extremely high recognition accuracy for various tasks such as general object recognition\cite{googlenet}, detection\cite{fast_rcnn}\cite{faster_rcnn} and semantic segmentation\cite{fcn}\cite{segnet}.
Since AlexNet\cite{alexnet} won the ImageNet Large Scale Visual Recognition Challenge (ILSVRC) 2012, which involved computation of 1,000 categories, network models with large numbers of layers began to appear, including VGG-16\cite{vggnet} and Residual Network (ResNet)\cite{resnet}, producing remarkable increases in speed and dropping error rates.
Recently, there is a continuing trend toward increasing recognition accuracy by increasing the number of layers, as was done with VGG-16 and ResNet.
As recognition accuracy increases in this way, models are becoming more complex, and increasing computation time and model size is becoming an issue.
Table 1 shows the amount of computation and model sizes for AlexNet, VGG-16 and ResNet, which are de facto standards for DCNN network models.
\begin{table}[t]
  \centering
  \caption{Comparison of DCNN and Binarized-DCNN}
  \label{tb:number_of_operations_and_model_size}
  {\tiny 
  \begin{tabular}{c|c|c|c|c|c|c} \hline
    \multirow{2}{*}{Model} & \multicolumn{4}{c|}{Computations [million]} & \multicolumn{2}{c}{Model size [MB]} \\ \cline{2-7}
                           & \multicolumn{2}{c|}{Convolutional} & \multicolumn{2}{c|}{Fully connected} & Convolutional & Fully connected      \\ \hline\hline
    \multicolumn{7}{c}{ImageNet classification task}                                       \\ \hline
    AlexNet   & \multicolumn{2}{c|}{ 10.77} & \multicolumn{2}{c|}{0.55}  &  14.29 & 237.91 \\ \hline
    VGG-16    & \multicolumn{2}{c|}{153.47} & \multicolumn{2}{c|}{1.26}  &  56.12 & 471.63 \\ \hline
    ResNet    & \multicolumn{2}{c|}{113.96} & \multicolumn{2}{c|}{0.021} & 223.90 &   7.81 \\ \hline
    Proposed  & AND      &  6.06            & AND      & 0.31            &        &        \\ \cline{2-5}
    (AlexNet) & bitcount &  6.06            & bitcount & 0.31            &   2.71 &  42.14 \\ \cline{2-5}
              & multiply &  0.24            & multiply & 0.003           &        &        \\ \hline
    Proposed  & AND      & 86.33            & AND      & 0.70            &        &        \\ \cline{2-5}
    (VGG-16)  & bitcount & 86.33            & bitcount & 0.70            &  10.62 &  88.64 \\ \cline{2-5}
              & multiply &  4.88            & multiply & 0.0033          &        &        \\ \hline
    Proposed  & AND      & 63.67            & AND      & 0.012           &        &        \\ \cline{2-5}
    (ResNet)  & bitcount & 63.67            & bitcount & 0.012           &  43.22 &   1.49 \\ \cline{2-5}
              & multiply &  7.93            & multiply & 0.00006         &        &        \\ \hline

    \multicolumn{7}{c}{Places205 scene recognition task}                                 \\ \hline
    AlexNet   & \multicolumn{2}{c|}{ 10.77} & \multicolumn{2}{c|}{0.55} & 14.29 & 211.20 \\ \hline
    VGG-16    & \multicolumn{2}{c|}{153.47} & \multicolumn{2}{c|}{1.20} & 56.12 & 459.20 \\ \hline
    Proposed  & AND      & 6.06             & AND       & 3.12          &       &        \\ \cline{2-5}
    (AlexNet) & bitcount & 6.06             & bitcount  & 3.12          &  2.71 & 39.79  \\ \cline{2-5}
              & multiply & 0.24             & multiply  & 0.003         &       &        \\ \hline
    Proposed  & AND      & 6.06             & AND       & 3.12          &       &        \\ \cline{2-5}
    (VGG-16)  & bitcount & 6.06             & bitcount  & 3.12          & 10.62 & 71.91  \\ \cline{2-5}
              & multiply & 0.24             & multiply  & 0.003         &       &        \\ \hline
  \end{tabular}
  }
\end{table}
The network model for AlexNet, which won ILSVRC 2012, is composed of five convolutional layers and three fully-connected layers.
The network model for VGG-16, which placed second in ILSVRC 2014, is composed of 13 convolutional layers and three fully-connected layers.
The network model for ResNet, which won ILSVRC 2015, is composed of 152 convolutional layers and one fully-connected layer.
Table 1 shows that convolutional layers accounted for more than 90\% of computation in AlexNet, and more than 99\% in VGG-16 and ResNet.
Although fully-connected layers accounted for roughly 90\% of the AlexNet and VGG-16 model sizes, ResNet had roughly the same model size as AlexNet, even though it had 152 layers. This shows that the convolutional layers contribute more to the amount of computation, and the fully connected layers contribute more to model size.
It is imperative to increase the speed of recognition processing and to decrease model size in order to use these methods in environments with limited resources, such as embedded and mobile devices.
To resolve this issue, research on accelerating computation and compressing model sizes has been proposed.

BinaryNet\cite{binary_net}, Binarized Neural Networks\cite{bdnn} and XNOR-Net\cite{xnor_net} are proposed methods to simultaneously accelerate processing and reduce memory use by binarizing DCNN.
Both networks express activation values and weights as binary values and express parameters as single bits to reduce memory size and perform high-speed inner products.
These methods are able to increase speed and effectively decrease memory use, as is needed, but they require retraining, so they cannot be applied to existing network models.
As such, our research proposes Binary-decomposed DCNN, which is able to accelerate inference computation and compress model size for existing network models, without requiring retraining.
Binary-decomposed DCNN accelerates recognition processing and compresses models for existing network models by converting feature maps and weightings, which are used in recognition computations in each layer, to binary values and using approximate inner-product computations.
The contributions of this method are as follows:
\begin{enumerate}
  \item Simultaneously accelerates recognition computation and compresses models without the need for re-training by using many binary values and a small number of real values to approximate real-valued parameters. 
  \item Converts real-valued feature maps to binary feature maps in real time by introducing a quantization sub-layer.
  \item Can be applied to large-scale network models without great loss of accuracy, unlike BinaryNet and XNOR-Net.
\end{enumerate}

\section{Related work}
VGG-16 and ResNet achieved high recognition performance in ILSVRC, but models with many layers are complex, so long computation times and large model sizes were issues. To use such models in environments with limited resources, such as embedded devices and smartphones, it will be essential to accelerate recognition processing and compress model sizes. Various research has proposed ways to accelerate processing and compress models, solving these issues.

\subsection{Compressing model size by eliminating parameters}
Deep Compression\cite{deep_compression} and SqueezeNet\cite{squeeze_net} are research on compressing model size.
Deep Compression combines branch pruning, quantization and Huffman encoding to compress model size by approximately 1/50, while increasing performance.
It first eliminates connections in a trained model by setting all weight below a certain threshold to zero.
This enables the dense weight matrix to be handled as a sparse matrix.
The model can then be effectively compressed further using storage methods such as Compressed Sparse Row (CSR) or Compressed Sparse Column (CSC).
Similar weights can also be shared by applying clustering to the sparse weight matrix.
Finally, the model is compressed using Huffman encoding.
The distribution of shared weight indices is uneven so this also improves efficiency of memory use.
SqueezeNet introduces Fire modules, compressing models by approximately 1/50 while achieving performance comparable to AlexNet.
A Fire module is composed of a Squeeze layer, which replaces $3 \times 3$ weight filters with $1 \times 1$ weight filters, and an Expand layer, which uses multiple $1 \times 1$ and $3 \times 3$ weight filters.
Using the Squeeze layer at the first stage reduces the dimensionality of the weight filters, and reduces the number of channels needed in the Expand layer.
SqueezeNet uses schemes such as introducing the Expand layer and down-sampling in the lower layers preserve inference performance.

\subsection{Acceleration and model compression using binary parameters}
BinaryNet\cite{binary_net}, Binarized Neural Networks\cite{bdnn} and XNOR-Net\cite{xnor_net} are methods that simultaneously accelerate computation and compress memory use by Binarizing DCNN feature maps and weights.
BinaryNet expresses activation and weight values as binary values, expressing parameters as single bits to reduce memory size, and enabling fast inner product computations.
BinaryConnect\cite{binary_connect} is used to binarize activation values and weights.
When updating parameters, real-valued parameters rather than the binarized parameters are updated.
Although binarizing activation and weight values achieves both faster computation and model compression, parameters cannot be updated through back-propagation of error.
As such, BinaryNet computes updated weights by replacing some parameters only when performing parameter clipping and gradient computation.
XNOR-Net improves on the accuracy of BinaryNet by introducing scaling coefficients.
To do so, both binary filters and scale factors are approximated to minimize the approximation error due to each weights in the BinaryConnect binarization method.
To compute the parameter updates, a method similar to BinaryNet is used.
There are also regions during convolution computations where inner product computations are duplicated.
This efficiency is improved by computing the average absolute value in the channel direction and convolving that output with the binary filter. 

\section{Proposed method}
Our proposed method simultaneously accelerates inference computation and compresses models for DCNN by transforming feature maps and weights to binary values.
The proposed method consists of two parts: (1) decomposing real-valued vector of weights to binary basis vectors, (2) quantization sub-layer.

\subsection{Decomposing real-valued vector to binary basis vectors}
To use binary inner product operations, real-valued parameters must be converted to binary values.
One method for converting parameters to binary is vector decomposition\cite{greedy_vector_decomposition}\cite{exhaustive_vector_decomposition}\cite{spade}.
Vector decomposition breaks down a weight vector, $\mathbf{w} \in \mathbb{R}^D$, into a binary basis matrix, $\mathbf{M} \in \{-1, 1\}^{D \times k}$, and a scaling coefficient vector, $\mathbf{c} \in \mathbb{R}^{k}$.
Here, $k$ is the number of basis vectors, or basis rank, and $D$ is the input dimensionality.
By applying vector decomposition to the weight vectors, inner products between two real values can be replaced with inner products between binary values when an input vector, $\mathbf{x}$, is binarized.
Inner products between binary values can be done quickly using logical operations and bit counting.
Two algorithms for optimizing vector decomposition are the greedy algorithm\cite{greedy_vector_decomposition} and the exhaustive algorithm\cite{exhaustive_vector_decomposition}.
In this section, we describe decomposition using an exhaustive optimization algorithm which is better than the greedy algorithm.

\subsubsection{Exhaustive algorithm \cite{exhaustive_vector_decomposition}}
Decomposition by the exhaustive algorithm computes a binary basis matrix, $\mathbf{M}$, and scaling vector, $\mathbf{c}$, that minimize the cost function in Eqn. \ref{eq:cost_function} on the weight vector, $\mathbf{w}$.
The decomposition is very slow, optimizing $\mathbf{M}$ through exhaustive search, but it can provide a decomposition with better approximation performance than the greedy algorithm.
The decomposition algorithm is shown in Algorithm \ref{alg:exhaustive_optimization}.
First, $\mathbf{M}$ is initialized to random values from $\{-1, 1\}$, and $\mathbf{c}$ is initialized to random real values.
Then, $\mathbf{M}$ and $\mathbf{c}$ are optimized.
It is difficult to optimize both $\mathbf{M}$ and $\mathbf{c}$ simultaneously, so they are each optimized alternately.
$\mathbf{M}$ is fixed, and $\mathbf{c}$ is optimized by minimizing Eqn. \ref{eq:cost_function} using the least squares method.
Then, $\mathbf{c}$ is fixed and $\mathbf{M}$ is optimized by exhaustive search. This process is repeated until the value of the cost function, Eqn. \ref{eq:cost_function}, converges.
Note that the accuracy of approximation of the vector decomposition depends on the initial values, so we take the values of $\mathbf{M}$ and $\mathbf{c}$ that minimize Eqn. \ref{eq:cost_function} after changing the initial values $L$ times as the basis decomposition result.
\begin{eqnarray}
  E = ||\mathbf{w} - \mathbf{Mc}||_2^2
  \label{eq:cost_function}
\end{eqnarray}

\begin{algorithm}[ht]
  \centering
  \caption{Decomposition algorithm}
  \label{alg:exhaustive_optimization}
  \begin{algorithmic}
    \REQUIRE $\mathbf{w}$, $k$, $L$
    \FOR{$i$ to $L$}
      \STATE Initialize $\mathbf{M}_i$ by random values on $\{-1, 1\}$
      \REPEAT
        \STATE $\mathbf{c}_i = (\mathbf{M}_i^{\mathrm{T}} \mathbf{M}_i)^{-1} (\mathbf{M}_i^{\mathrm{T}} \mathbf{w})$
        \STATE $\mathbf{M}_i = \argmin_{\mathbf{M}_i \in \{-1, 1\}^{D \times k}} ||\mathbf{w} - \mathbf{M}_i \mathbf{c}_i ||_2^2$
      \UNTIL{$||\mathbf{w} - \mathbf{Mc}||_2^2$ converges}
      \STATE  $\hat{\mathbf{M}}, \hat{\mathbf{c}} = \argmin_{\mathbf{M, c}} ||\mathbf{w - Mc}||_2^2$
    \ENDFOR
    \RETURN $\hat{\mathbf{M}}, \hat{\mathbf{c}}$
  \end{algorithmic}
\end{algorithm}

\subsubsection{Applying vector decomposition to the convolution layers}
First, consider application of vector decomposition to the $m$th weight filter associated with the $n$th feature map in the $l$th layer, $\mathbf{w}_{n,m}^l \in \mathbb{R}^{H \times H}$.
Vector decomposition applies to vectors, so it cannot be applied directly to the weight filter, which is a matrix.
As such, we apply vector decomposition by expressing the weight filter as a vector.
Expressing the weight filter as a vector results in a vector of dimension $H \cdot H$.
With network models as in VGG-16 and ResNet, the weight filter for each layer is usually very small, so this does not reduce the amount of computation using approximate inner-product computations has little effect.
Thus, rather than decomposing a single weight filter, filters in the channel direction are expressed as a vector.
With $\mathbf{M}$ as the number of input feature maps, the decomposed weight vectors can be defined as .
Then, the dimension of $\mathbf{W}_n$ is $M\cdot H^2$, so approximate inner-product computations have more effect.
In convolutional layers, $\mathbf{W}$ only has $N$ output maps.
Vector decomposition using Algorithm \ref{alg:exhaustive_optimization} is applied to each $\mathbf{W}$, decomposing them into  and . In convolutional layer inference processing, these  and  are used for approximate inner product calculations.

\subsubsection{Applying vector decomposition to the fully-connected layers}
Next, consider application of vector decomposition to connection weights in the $n$th unit of the $l$th fully-connected layer, $\mathbf{w}_n^l \in \mathbb{R}^M$.
The weights used to get the output from the $n$th unit of a fully connected layer are an $M$-dimensional weight vector.
Also a fully connected layer has values equaling the number of output units, $N$.
Vector decomposition using Algorithm \ref{alg:exhaustive_optimization} is applied to each weight vector, $\mathbf{w}_n^l$, to decompose into the binary basis matrix, $\hat{\mathbf{M}}$, and the scaling coefficient vectors, $\hat{\mathbf{c}}_n$.

\begin{figure*}[th]
  \centering
  \includegraphics[width=160mm]{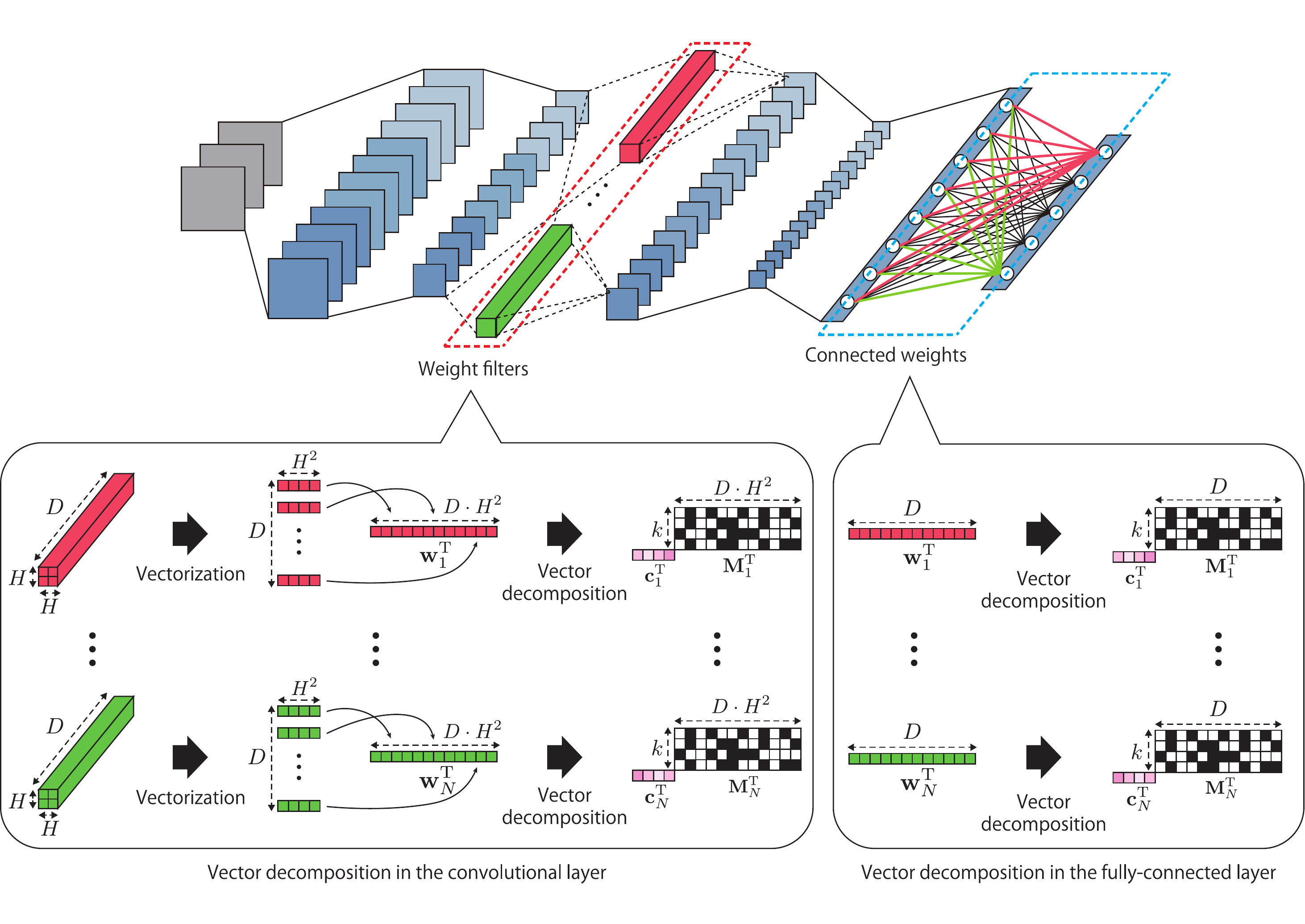}
  \caption{Weights decomposition in DCNN layers}
  \label{fig:decomposition_for_dcnn}
\end{figure*}

\subsection{Quantization sub-layer}
Quantization can convert to binary rapidly, but it applies to a fixed range, so real and negative values cannot be quantized.
Thus, we introduce a quantization sub-layer able to binarize real values, including negative values, rapidly.
The quantization sub-layer is able to quantize real values including negative values by changing the quantization range. 

Before quantizing a feature map, the quantization bit-depth, $Q$, is decided.
The approximation accuracy of quantization increases with larger $Q$, but the amount of computation required for approximate inner products also increases, making computation slower.
Conversely, as $Q$ becomes smaller, the amount of computation decreases, so inference computations become faster, but approximation accuracy decreases due to quantization.
First, Eqn. (2) is used to find the quantization range, $\Delta d$，between the maximum and minimum values in feature map.
$\Delta d$ depends on the maximum and minimum values in the feature map, so the value is different for each feature map.

\begin{eqnarray}
  \Delta d = \frac{\max{(\mathbf{x})} - \min{(\mathbf{x})}}{2^Q - 1}
  \label{eq:calc_quantization_level}
\end{eqnarray}

Next, Eqn. (\ref{eq:shift_feature_maps}) is used to shift the minimum value of the feature map to $0$.
Here, $\mathbf{1}$ represents the unit vector.
Shifting the feature map enables quantization of even negative values, which normally could not be quantized. 
\begin{eqnarray}
  \mathbf{x}' = \frac{\mathbf{x} - \mathbf{1}\min(\mathbf{x})}{\mathbf{1}Q}
  \label{eq:shift_feature_maps}
\end{eqnarray}

Finally, $\mathbf{x}'$ is quantized.
Quantizing $\mathbf{x}'$ generates a binary code, $\mathbf{B}\in\{0,1\}^{D\times Q}$. 
The binary code from the quantization sub-layer can be recovered using Eqn. \ref{eq:approx_feature_maps}.
\begin{eqnarray}
  \mathbf{x} \approx \mathbf{Br} + \mathbf{1}\min(\mathbf{x})
  \label{eq:approx_feature_maps}
\end{eqnarray}


\subsection{Inference processing}
Binary-decomposed DCNN accelerates forward computation of network using approximate binary inner product computations.
To perform operations on two binary values, the quantization sub-layer is introduced to binarize feature maps $\mathbf{B}$.
Feature map values and feature vectors in each layer depend on the input samples, so in contrast to vector decomposition of the weights, quantization cannot be done ahead of time.
The ability of the quantization sub-layer to rapidly binarize real values is used to perform this quantization during inference computations.

\subsubsection{Computation in convolution layers}
\label{inference_convolution}
For ordinary convolution computations, the $n$-th feature map $u_{n,i,j}$ is obtained as the sum of products of the local area of feature mapping $\mathbf{x}_{i,j}$, and weight filter $\mathbf{w}^\mathrm{T}_n$.
In our method the output is computed by replacing the sum of products of real values with binary operations.
Input feature maps are first quantized by the quantization sub-layer.
This generates binary feature maps, $\mathbf{B}\in\{0,1\}^{MH^2\times Q}$.
Then the convolution is computed using the binary feature maps and binary weight filters, $\mathbf{M}_n^\mathrm{T}$ and $\mathbf{c}_{n}^\mathrm{T}$, obtained through vector decomposition.
The output $u_{i,j,n}$ is given by Eqn. \ref{eq:inference_convolution}.
\begin{eqnarray}
  u_{n,i,j} &=      & \mathbf{w}_n^ \mathrm{T} \mathbf{x}_{i,j} \nonumber \\
            &\approx& \hat{\mathbf{c}}_n^\mathrm{T} \hat{\mathbf{M}}_n^\mathrm{T} (\mathbf{B}_{i,j} \mathbf{r}_{i,j} + \mathbf{1} \min(\mathbf{x})) \nonumber \\
            &=      & \hat{\mathbf{c}}_n^\mathrm{T} \hat{\mathbf{M}}_n^\mathrm{T} \mathbf{B}_{i,j} \mathbf{r}_{i,j} + \hat{\mathbf{c}}_n^\mathrm{T} \hat{\mathbf{M}}_n^\mathrm{T} \mathbf{1}\min(\mathbf{x})
  \label{eq:inference_convolution}
\end{eqnarray}
Here, $\mathbf{w}_n^\mathrm{T}\in\mathbb{R}^{DH^2}$ is the weighting used when generating the $n$-th feature map, and $\mathbf{x}_{i,j}\in\mathbb{R}^{DH^2}$ is the feature map used when generating the unit at coordinates $i,j$ in the output feature map.

\subsubsection{Computation in fully-connected layers}
The $i$-th output $u_i$, from fully-connected layers are computed by inner products between real-valued feature vectors, $\mathbf{x}$, and connection weights, $\mathbf{w}_i$.
To accelerate inner-product calculations in fully-connected layers, real values are replaced with binary values, as in the convolutional layers.
This generates binary feature matrix $\mathbf{B} \in \{0,1\}^{D\times Q}$ from the input feature vector $\mathbf{x}\in \mathbb{R}^D$.
Then, the output $i$ is approximated by Eqn. \ref{eq:calc_approx_fully_connected} from the binary feature matrix and the binary weight vectors decomposed beforehand. 
\begin{eqnarray}
  u_i &   =   & \mathbf{w}_i^\mathrm{T} \mathbf{x} \nonumber \\
      &\approx& \hat{\mathbf{c}}_i^\mathrm{T} \hat{\mathbf{M}}_i^\mathrm{T} (\mathbf{B} \mathbf{r} + \mathbf{1}\min(\mathbf{x})) \nonumber \\
      &   =   & \hat{\mathbf{c}}_i^\mathrm{T} \hat{\mathbf{M}}_i^\mathrm{T} \mathbf{B} \mathbf{r} + \hat{\mathbf{c}}_i^\mathrm{T} \hat{\mathbf{M}}_i^\mathrm{T} \mathbf{1}\min(\mathbf{x})
  \label{eq:calc_approx_fully_connected}
\end{eqnarray}
where, $\hat{\mathbf{M}}^\mathrm{T} \in \{-1, 1\}^{k\times D}$ and $\mathbf{B} \in \{0, 1\}^{D\times Q}$ are binary, so it can be computed using logical operators and bit counting, as in Eqn. \ref{eq:binary_inner_product}.
The computation can be done at high speed, counting bits using the POPCNT function implemented in the Streaming SIMD Extension (SSE) 4.2.
\begin{eqnarray}
  \hat{\mathbf{m}}_i^\mathrm{T} \mathbf{b}_j = 2 \times \mathrm{POPCNT}(\mathrm{AND}(\hat{\mathbf{m}}_i^\mathrm{T}, \mathbf{b}_j)) - \|\mathbf{b}_j\|_1^2
  \label{eq:binary_inner_product}
\end{eqnarray}

\section{Experiments}
In testing, we evaluated recognition performance, processing time and model size when applying the proposed method to several network models.
Quantization bit depth, $Q$, of 4, 6 and 8 were used for approximations, and similarly, basis rank of 4, 6, and 8 were used.
Quantization bit depths and basis rank less than 4 were not used because error rates increased to a great degree.
Similarly, with quantization bit rate and basis rank greater than 8, the drop in error rate had peaked, so they were not included.
To evaluate model size in testing, the total memory occupied by the network model, including weights, $\mathbf{W}$, binarized basis matrix, $\mathbf{M}$, and scaling coefficient vector, $\mathbf{c}$, were compared.
Published trained models were used as parameters for each network model, with no fine tuning.
Top-5 accuracy was used to evaluate recognition performance.
Top-5 accuracy is a method that counts cases where the training signal is included among the five most probable inferred classes, as success.
We used an Intel Core i7-4770 3.40-GHz processor.

\subsection{ImageNet classification task}
Testing was done using the AlexNet, VGG-16, and ResNet-152 network models.
The ImageNet data set used in the ILSVRC 2012 \cite{ilsvrc} classification task was used.
ImageNet is a very large object recognition dataset, containing 1,200,000 training samples, 100,000 test samples, and 50,000 validation samples.
Each sample is classified into one of 1000 categories.
In testing, evaluation was done using the 50,000 validation samples.

\subsubsection{Model 1: AlexNet}
\label{alexnet_experiments}

\begin{figure*}[h]
  \subfigure[error vs acceleration]{
    \label{fig:alexnet_error_vs_acceleration}
    \includegraphics[width=50mm]{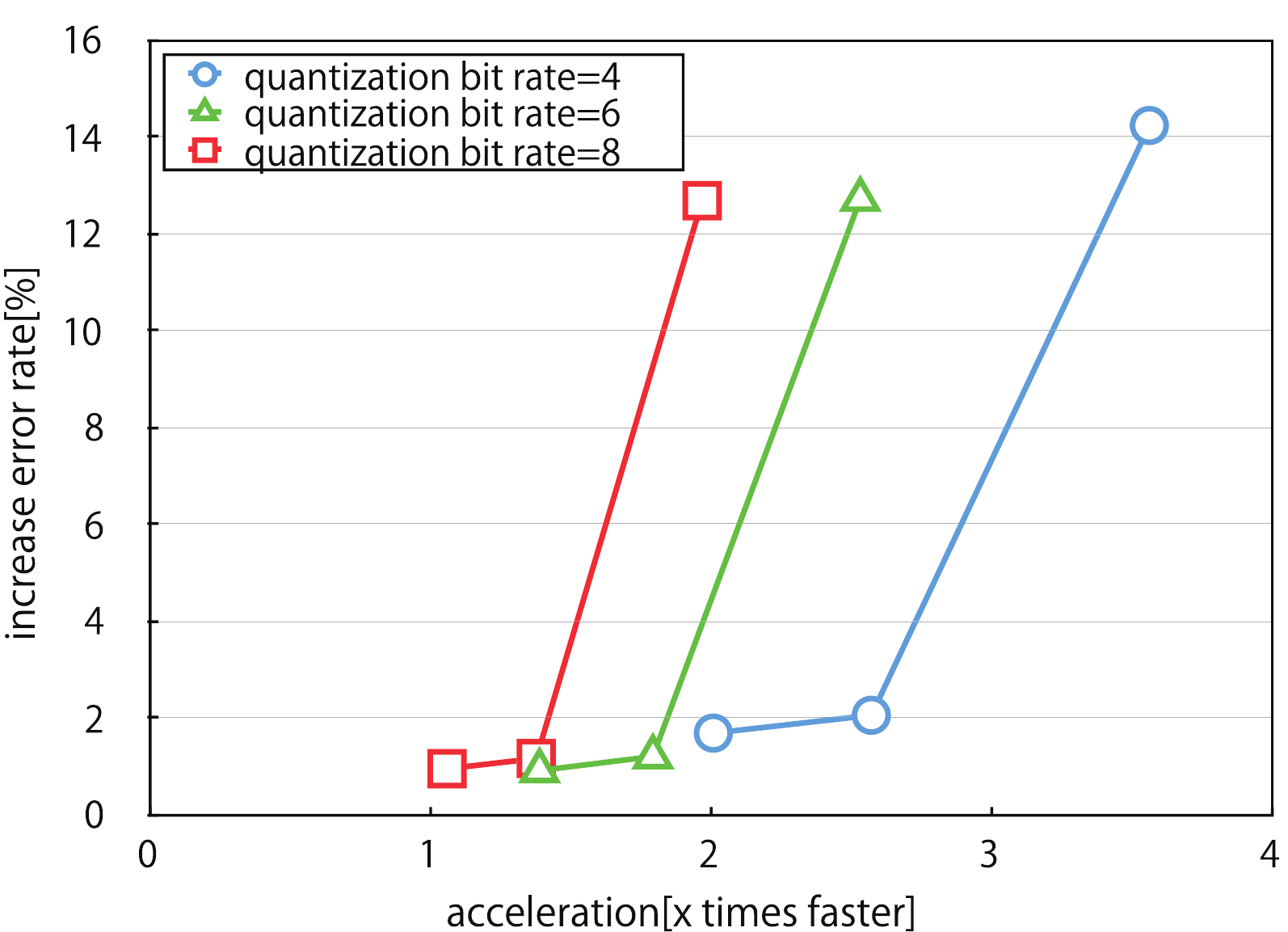}}
  \subfigure[error vs memory compression]{
    \label{fig:alexnet_error_vs_memory_compression}
    \includegraphics[width=50mm]{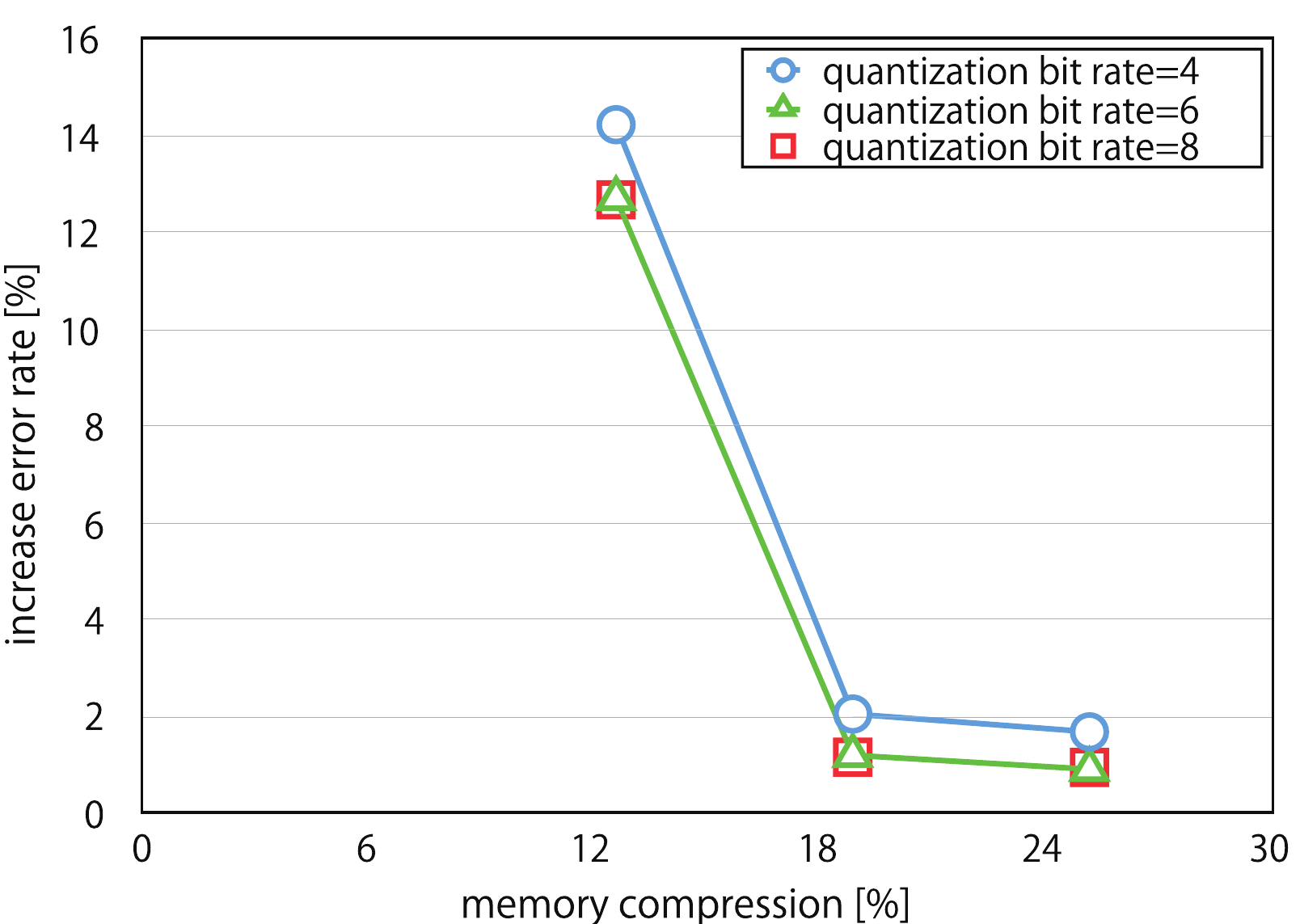}}
  \subfigure[process time]{
    \label{fig:alexnet_process_time}
    \includegraphics[width=50mm]{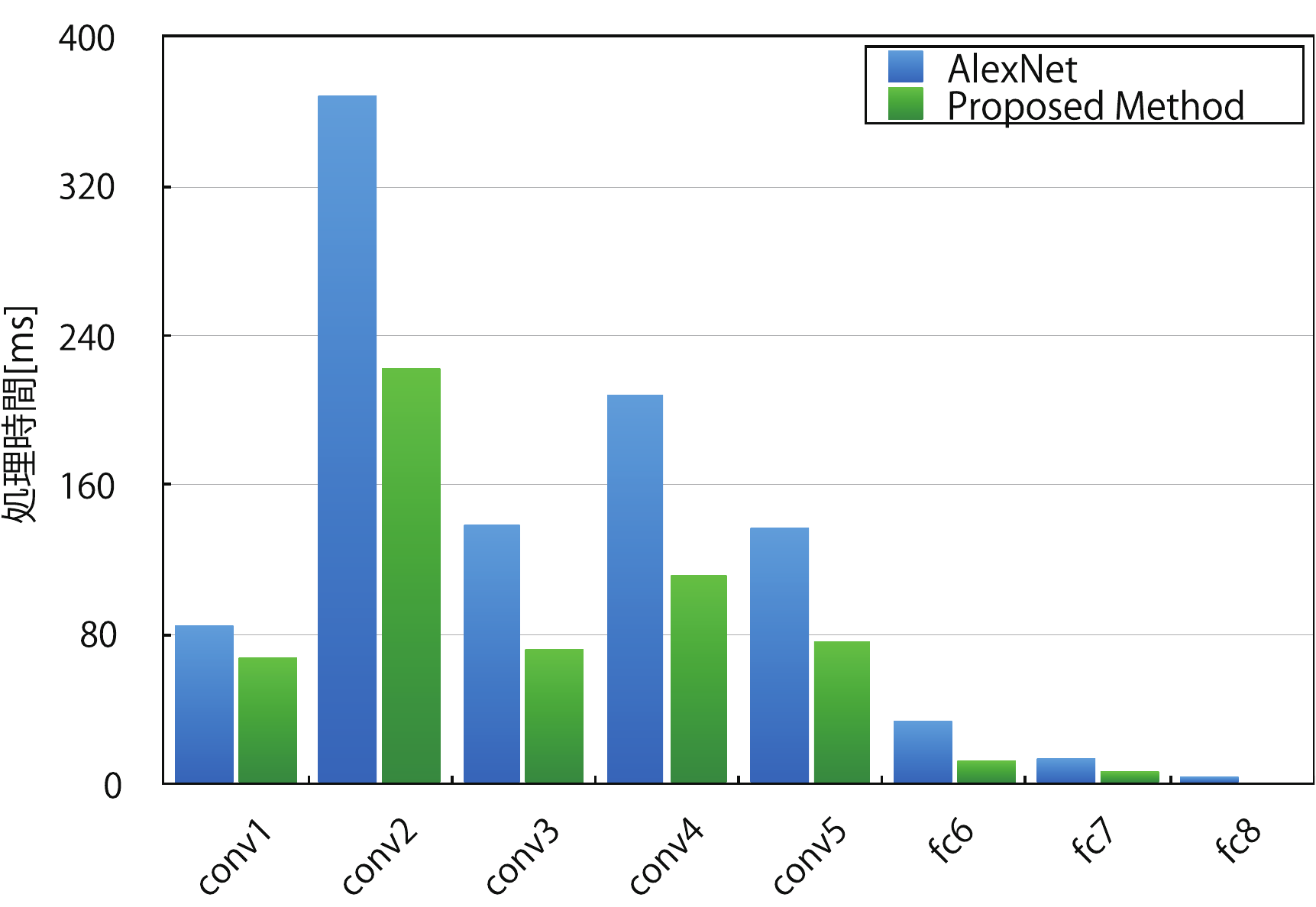}}
  \caption{ImageNet performance evaluation for AlexNet}
  \label{fig:alexnet_performance}

  \subfigure[error vs acceleration]{
    \label{fig:vgg16_error_vs_acceleration}
    \includegraphics[width=50mm]{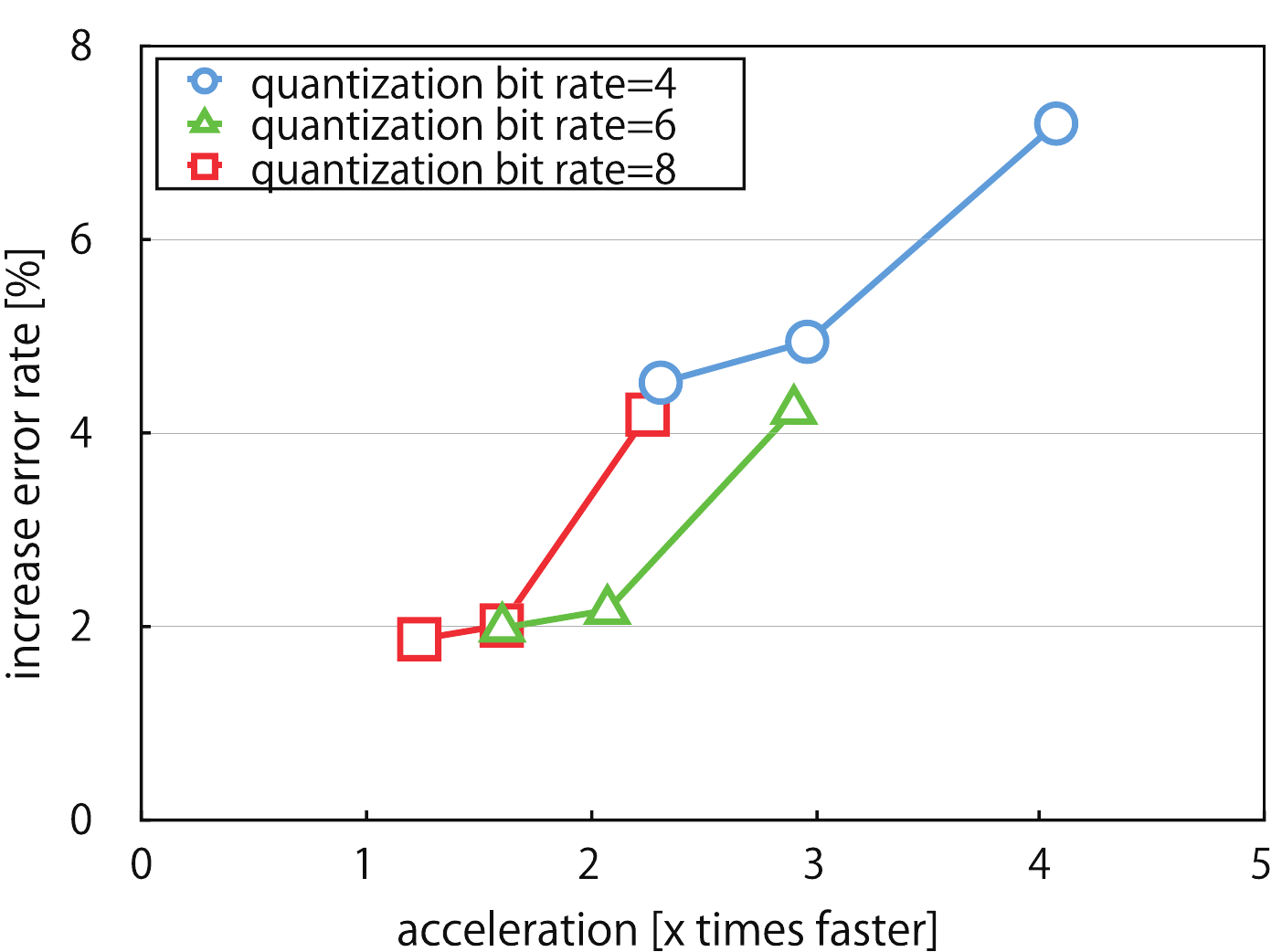}}
  \subfigure[error vs memory compression]{
    \label{fig:vgg16_error_vs_memory_compression}
    \includegraphics[width=50mm]{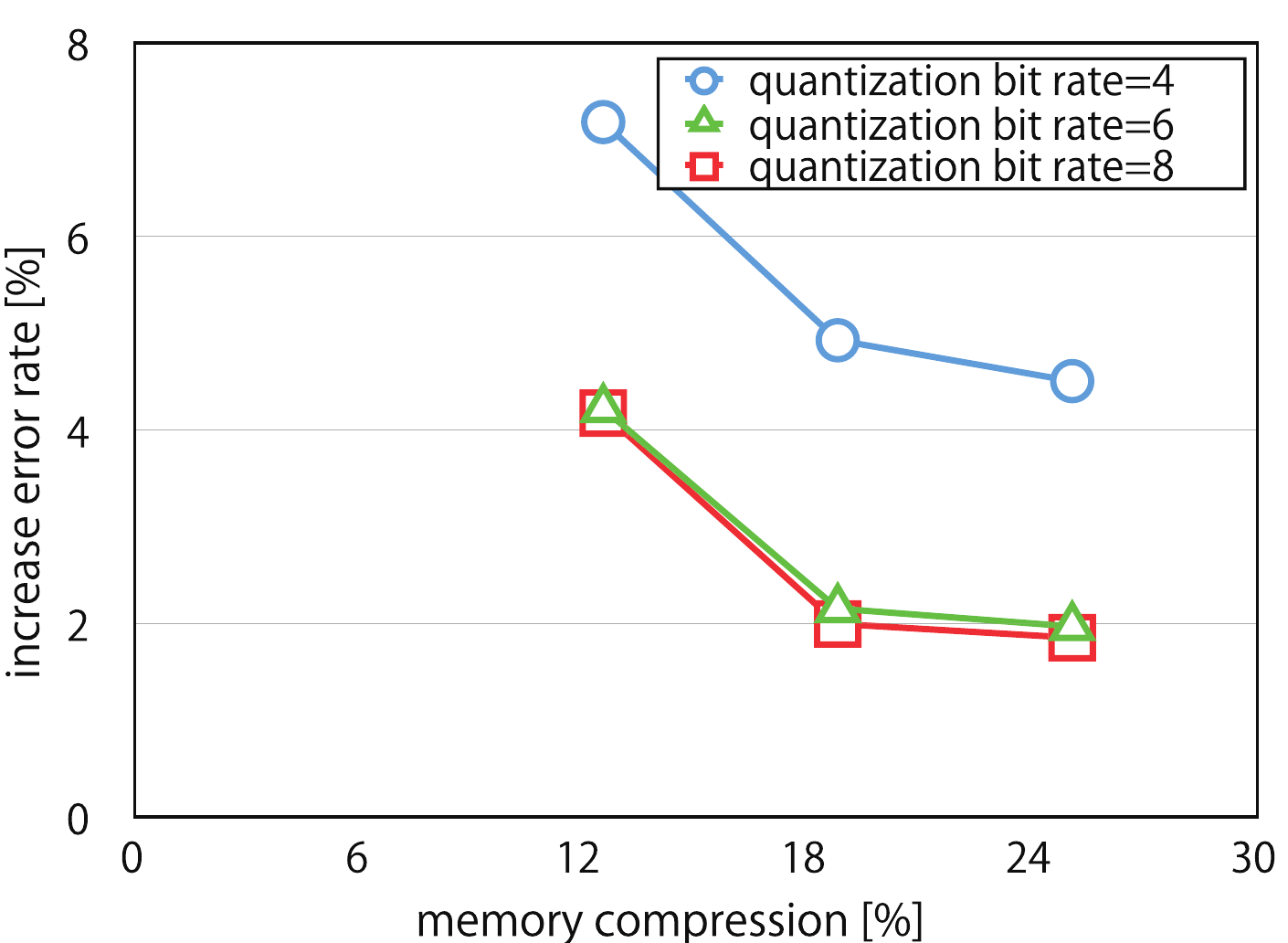}}
  \subfigure[process time]{
    \label{fig:vgg16_process_time}
    \includegraphics[width=50mm]{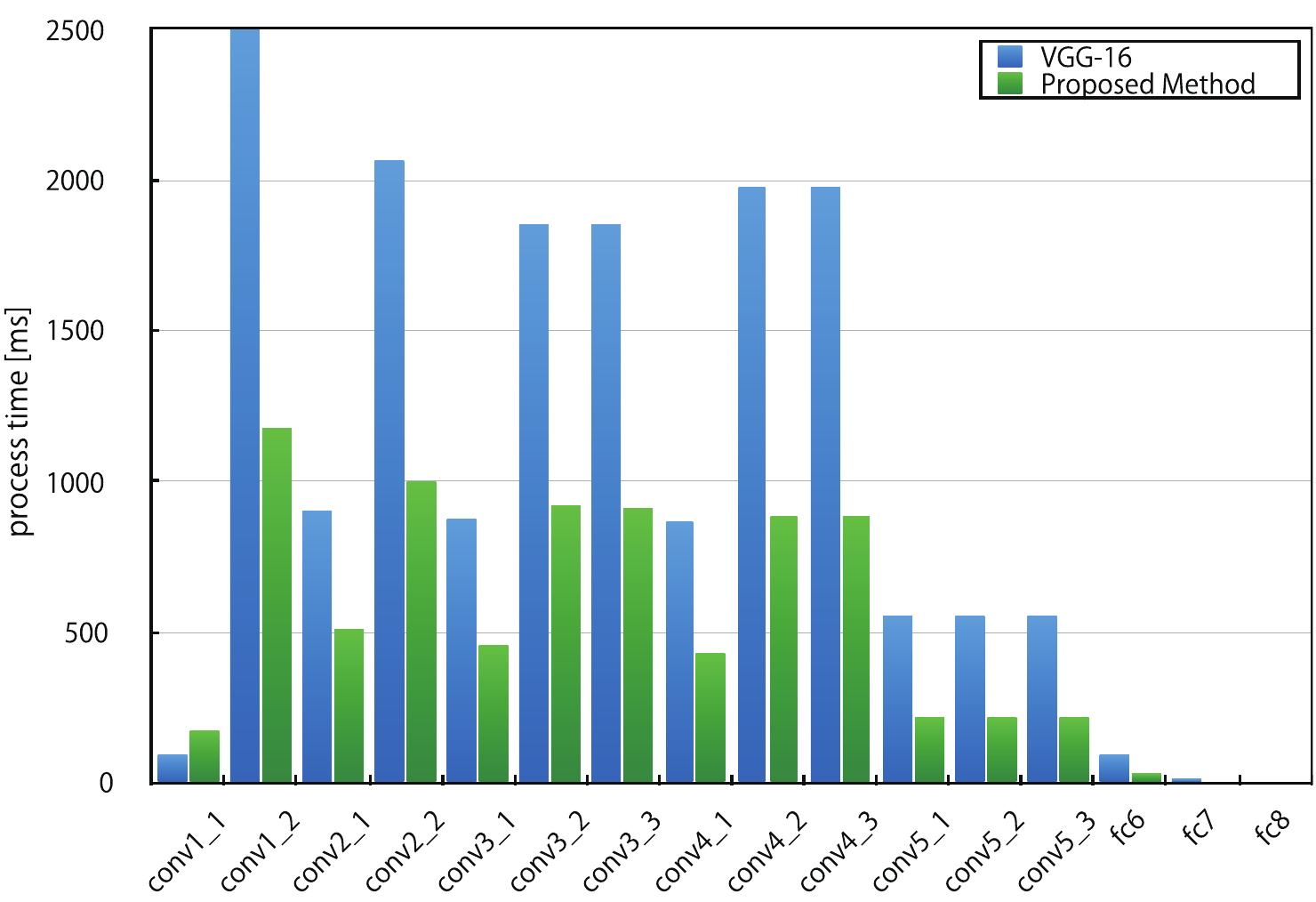}}
  \caption{ImageNet performance evaluation for VGG-16}
  \label{fig:vgg16_performance}

  \subfigure[error vs acceleration]{
    \label{fig:resnet152_error_vs_acceleration}
    \includegraphics[width=50mm]{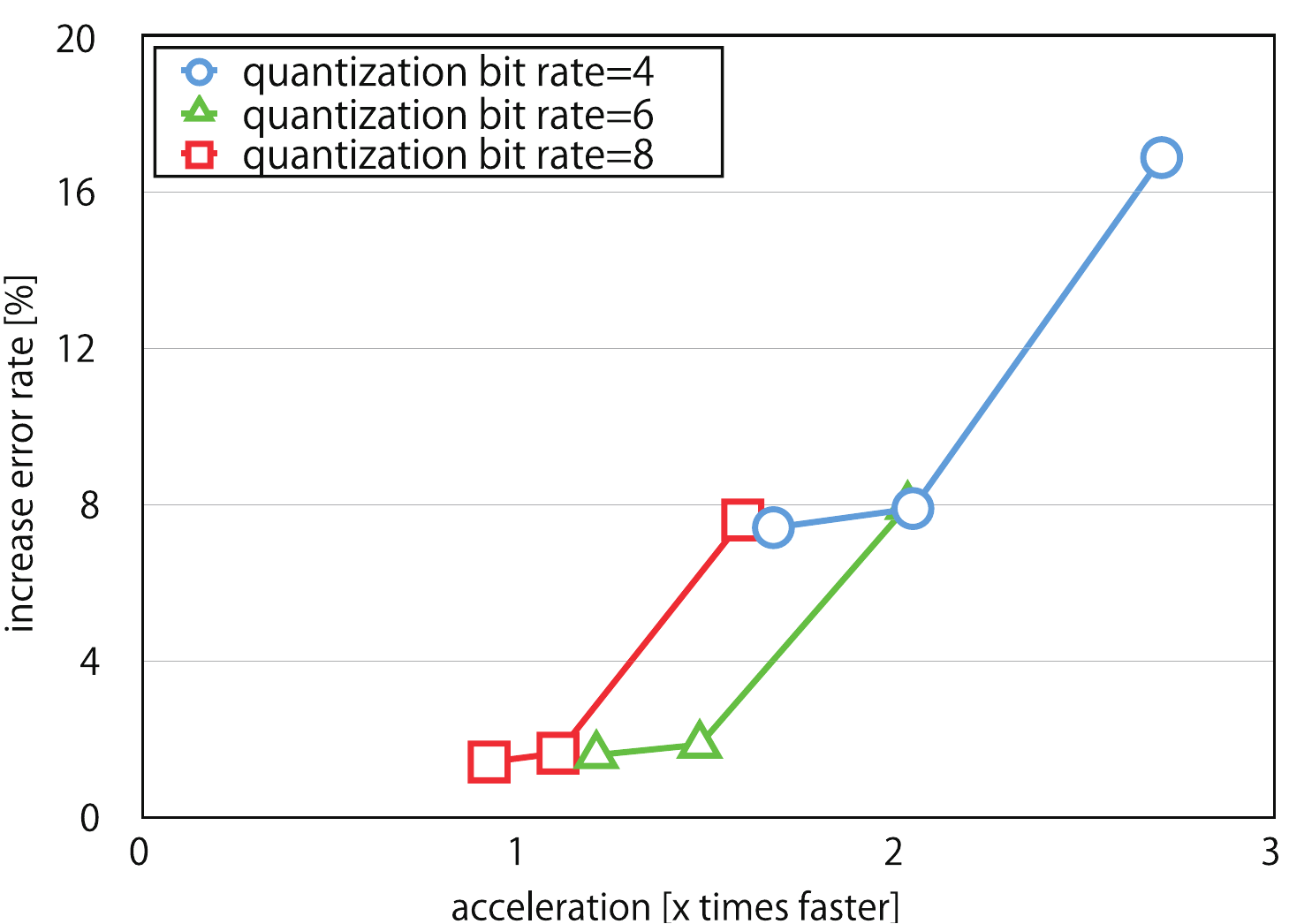}}
  \subfigure[error vs memory compression]{
    \label{fig:resnet152_error_vs_memory_compression}
    \includegraphics[width=50mm]{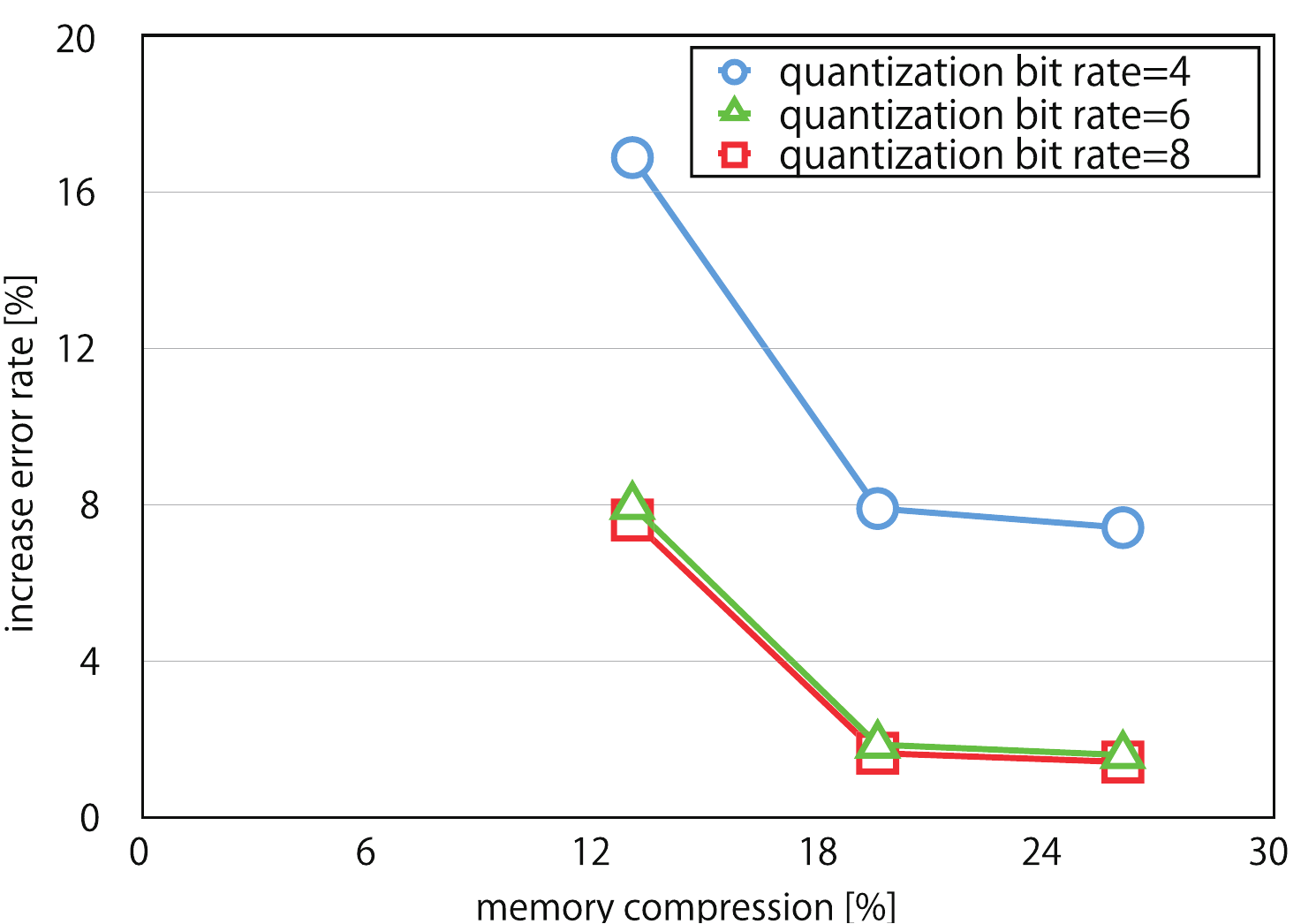}}
  \subfigure[process time]{
    \label{fig:resnet152_process_time}
    \includegraphics[width=50mm]{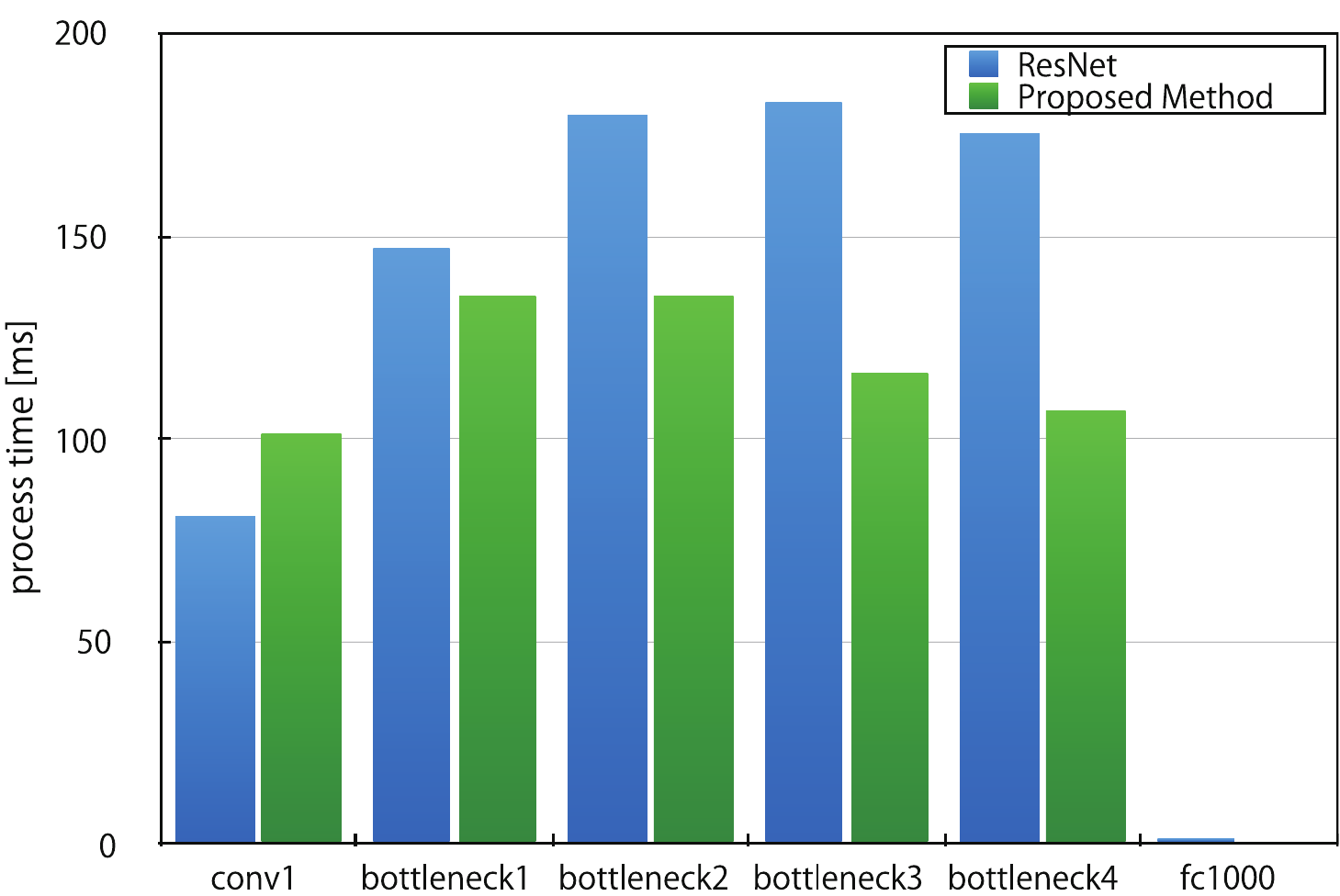}}
  \caption{ImageNet performance evaluation for ResNet-152}
  \label{fig:resnet152_performance}
\end{figure*}

AlexNet is composed of 5 convolutional layers and 3 fully-connected layers.
A comparison of recognition accuracy, processing time, and error-rate increases when using approximate inner-product calculations is shown in Figure \ref{fig:alexnet_performance}.
Here, $k$ indicates the basis rank used for weight decomposition.
Comparing the same basis rank for quantization bit-depths of 6 and 8, almost no increase in the error rate is shown.
In this case, the lower quantization bit-depth of 6 is better, requiring less computation.
Figure \ref{fig:alexnet_error_vs_memory_compression} shows that memory compression does not change between quantization bit depths of 4 and 6 using the same basis rank, so we can say quantization bit-depth does not affect reduction of memory use.
The error rate also does not change greatly when comparing basis ranks of 6 and 8.
This is similar to the trend in Figure \ref{fig:alexnet_error_vs_acceleration}.
For both quantization bit-depth and basis rank of 6, speed increased by a factor of 1.79, and model size decreased from 237.91 MB to 44.85 MB.
Here, error rates increased by 1.20\%.

\label{imagenet_experiments}

\subsubsection{Model 2: VGG-16}
\label{vgg16_experiments}
VGG-16 is composed of 13 convolutional layers and 3 fully connected layers.
The dimensions of the weight filter and number of input feature maps for the first convolutional layer are very small, so approximating inner-product computations may not produce significant effect.
For this reason, approximations were introduced on all but the first layer for testing.
Recognition accuracy, processing time, and model size with approximate inner-product computations are shown in Figure \ref{fig:vgg16_performance}.
With a quantization bit-depth of 6 and basis rank of 6, speed increased by a factor of 2.07 and the model size reduced from 527.74 MB to 99.26 MB.
In this case, the error rate increased by 2.16\%.

\subsubsection{Model 3: ResNet-152}
ResNet-152 is composed of 151 convolutional layers and 1 fully connected layer.
Recognition accuracy, processing speed and model size using the approximate inner product calculations are shown in Figure \ref{fig:resnet152_performance}.
ResNet has 152 convolutional layers, which account for approximately 96\% of the model size.
Clearly, reduction in model size can also be gained for models like ResNet, which have very many convolutional layers.
With a quantization bit-depth of 6 and basis rank of 6, speed increased by a factor of 1.77 and the model size reduced from 229.08 MB to 44.71 MB.
In this case, the error rate increased by 1.86\%.

\begin{table}[ht]
  \centering
  \caption{Comparison of performance with methods based on AlexNet}
  \label{tb:comparison_of_alexnet_based_models}
  {\footnotesize
  \begin{tabular}{c||c|c|c|c} \hline
    Model            & Top1 & Top5 & acceleration & compression \\ \hline\hline
    AlexNet          & 56.8 & 80.0 & -    & -  \\ \hline
    Deep Compression & 56.8 & 79.9 & 1.00 & 35 \\ \hline
    XNOR-Net         & 44.2 & 69.2 & 58.0 & 32 \\ \hline
    proposed         & 55.1 & 78.8 & 1.79 &  5 \\ \hline
  \end{tabular}
  }
\end{table}

\begin{figure*}[ht]
  \centering
  \subfigure[error vs acceleration]{
    \label{fig:segnet_error_vs_acceleration}
    \includegraphics[width=50mm]{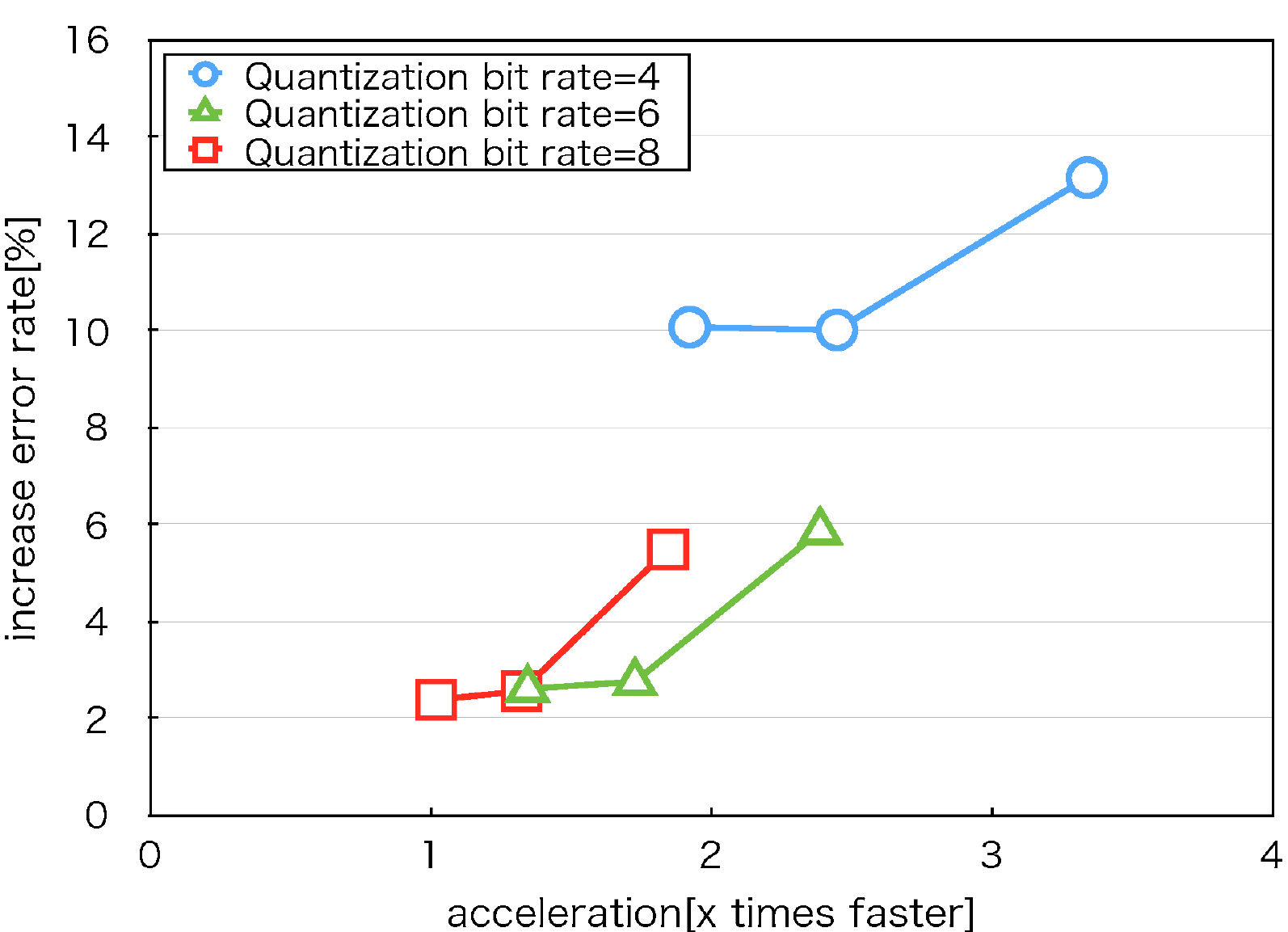}}
  \subfigure[error vs memory compression]{
    \label{fig:segnet_error_vs_memory_compression}
    \includegraphics[width=50mm]{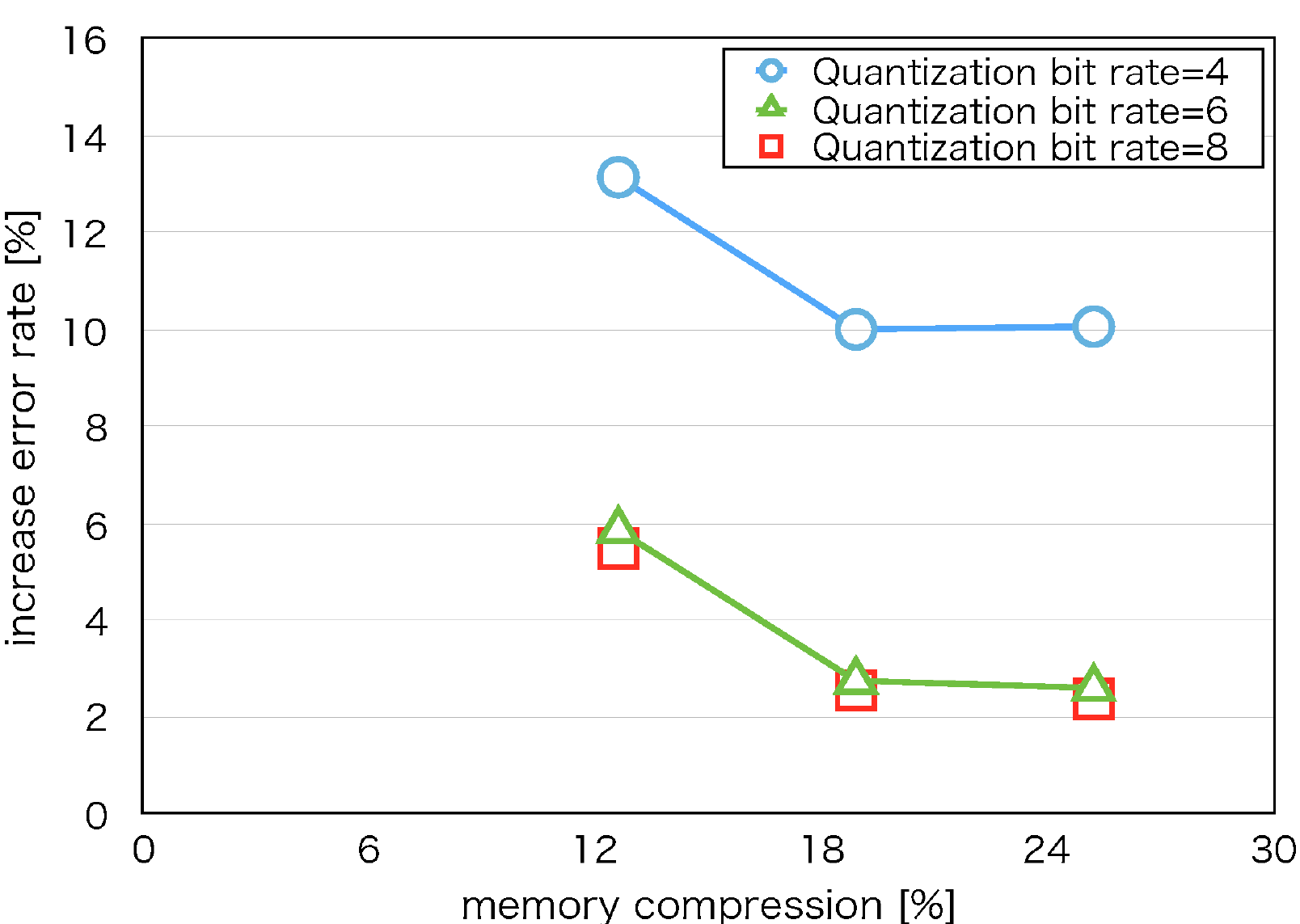}}
  \subfigure[process time]{
    \label{fig:segnet_process_time}
    \includegraphics[width=50mm]{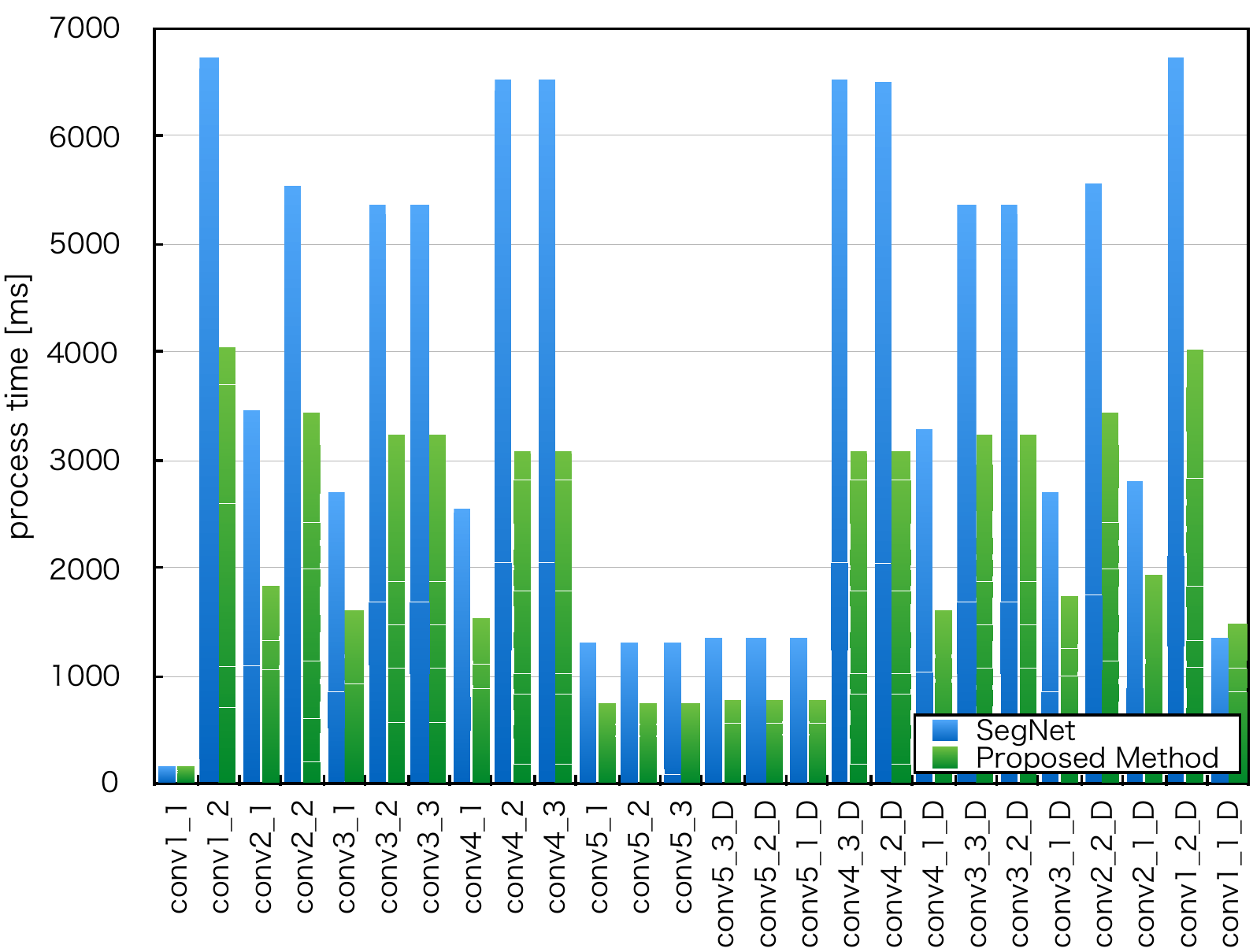}}
  \caption{Cityscapes dataset performance evaluation for SegNet}
  \label{fig:segnet_performance}
\end{figure*}
\begin{figure*}[ht]
  \centering
  \includegraphics[width=150mm]{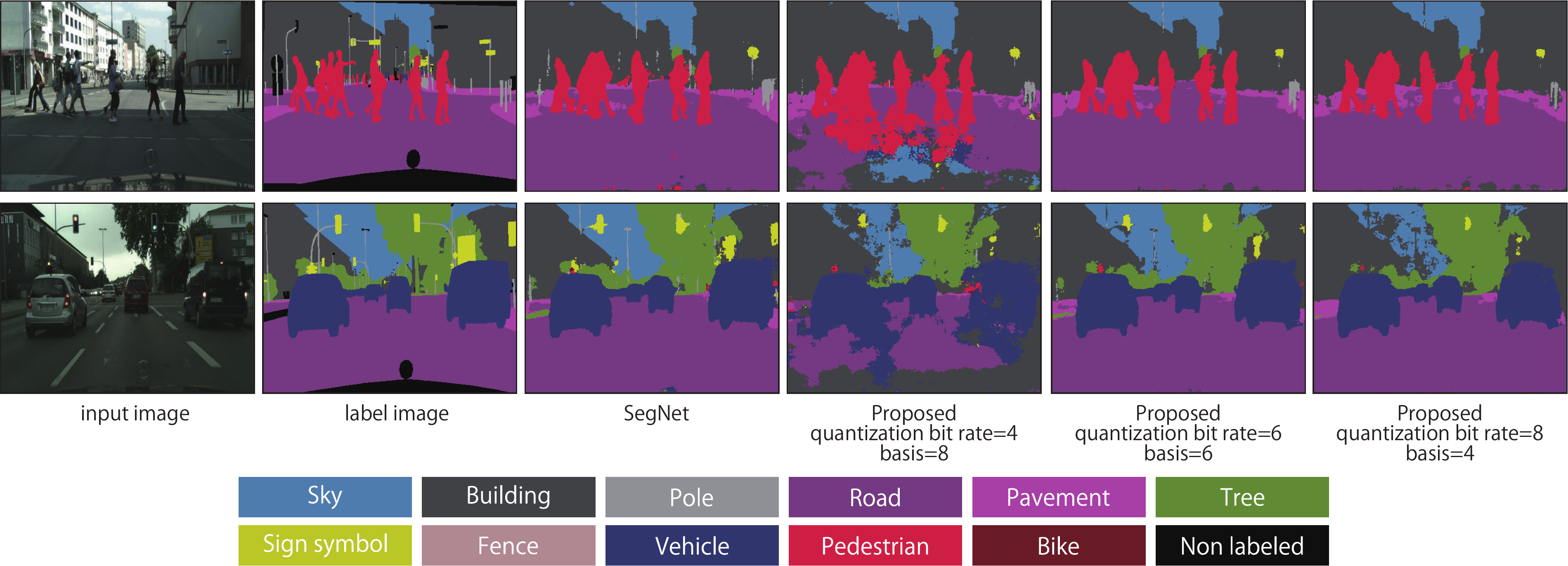}
  \caption{Semantic segmentation results in Cityscapes}
  \label{fig:segnet_visualize}
\end{figure*}

\subsubsection{Comparison with other methods}
In this section, we compare the top-1/5 accuracy, acceleration scale factors and model compression ratios of the proposed method with those of three other methods (AlexNet, Deep Compression and XNOR-Net).
The proposed method involves applying vector decomposition to a model with a quantization bit depth of 6. The comparison results are shown in Table \ref{tb:comparison_of_alexnet_based_models}.
Although Deep Compression had excellent compression performance, its recognition computation was no faster.
XNOR-Net had superior acceleration scaling performance and compression performance, but its recognition accuracy was much lower.
The proposed method had worse acceleration scaling and compression performance than the other methods, but maintained its recognition accuracy while simultaneously achieving a high acceleration scale factor and model compression.

\subsection{Cityscapes semantic segmentation task}
We performed an evaluation of semantic segmentation using the Cityscapes Dataset\cite{cityscapes}.
The Cityscapes Dataset is a very large segmentation dataset.
In this experiment, we performed the evaluation using the 500 verification samples.
For the network model, we used SegNet\cite{segnet}.
For the model parameters, we used published learned parameters, and fine tuning was not performed.
Figures \ref{fig:segnet_performance} and \ref{fig:segnet_visualize} show the recognition performance when using Cityscapes.
With a quantization bit depth of 6 and a basis rank of 6, we achieved an acceleration scale factor of approximately 1.73 and compressed the model size from approximately 112.25 MB to approximately 21.23 MB.
In this case, the rate of error increase was approximately 2.75\%.

\section{Conclusion}
We have proposed a method able to accelerate inference computation while compressing model sizes, using existing network models without the need for retraining.
The method compresses memory use by converting weightings from each layer from real-valued parameters to binary parameters, and accelerates inference computation by replacing real-valued inner product calculations with binary valued inner product calculations using logical operations and bit counting.
Using a quantization bit-depth of 6 and basis rank of 6, AlexNet model sizes were reduced by approximately 80\%, and speed increased by a factor of 1.79.
In this case, error rates increased by 1.20\%.
With VGG-16, model sizes reduced by 81\%, and speed increased by a factor of 2.07.
In this case, error rates increased by 2.16\%.
In the future, we plan to increase approximation accuracy and reduce the increases in error rates.

{\small
\bibliographystyle{ieee}
\bibliography{egbib}

\begin{thebibliography}{10}\itemsep=-1pt

\bibitem{spade}
M.~Ambai and I.~Sato.
\newblock {SPADE: Scalar Product Accelerator by Integer Decomposition for
  Object Detection}.
\newblock In {\em ECCV}, 2014.

\bibitem{squeeze_net}
F.~N. Candela, M.~W. Moskewicz, K.~Ashram, S.~Han, W.~J. Dally, and K.~Keutzer.
\newblock {SqueezeNet: AlexNet-level accuracy with 50x fewer parameters and
  $<$1MB model size}.
\newblock {\em arXiv preprint arXiv:1602.02830}, 2016.

\bibitem{cityscapes}
M.~Cordts, M.~Omran, S.~Ramos, T.~Rehfeld, M.~Enzweiler, R.~Benenson,
  U.~Franke, S.~Roth, and B.~Schiele.
\newblock The cityscapes dataset for semantic urban scene understanding.
\newblock In {\em Proceedings of the IEEE Conference on Computer Vision and
  Pattern Recognition (CVPR)}, 2016.

\bibitem{binary_net}
M.~Courbariaux and Y.~Bengio.
\newblock {BinaryNet: Training Deep Neural Networks with Weights and
  Activations Constrained to +1 or -1}.
\newblock {\em arXiv preprint arXiv:1602.02830}, 2016.

\bibitem{binary_connect}
M.~Courbariaux, Y.~Bengio, and J.~David.
\newblock {BinaryConnect: Training Deep Neural Networks with binary weights
  during propagations}.
\newblock {\em arXiv preprint arXiv:1511.00363}, 2015.

\bibitem{fast_rcnn}
R.~Girshick.
\newblock {Fast R-CNN}.
\newblock In {\em Proceedings of the International Conference on Computer
  Vision (ICCV)}, 2015.

\bibitem{deep_compression}
S.~Han, H.~Mao, and W.~J. Dally.
\newblock {Deep Compression: Compressing Deep Neural Networks with Pruning,
  Trained Quantization and Huffman Coding}.
\newblock In {\em Proceedings of the International Conference on Learning
  Representations (ICLR)}, 2016.

\bibitem{resnet}
K.~He, X.~Zhang, S.~Ren, and J.~Sun.
\newblock {Deep Residual Learning for Image Recognition}.
\newblock In {\em The IEEE Conference on Computer Vision and Pattern
  Recognition (CVPR)}, 2016.

\bibitem{bdnn}
I.~Hubara, D.~Soudry, and R.~E. Yaniv.
\newblock {Binarized Neural Networks}.
\newblock {\em arXiv preprint aiXiv:1602.02505}, 2016.

\bibitem{fcn}
L.~Jonathan, S.~Evan, and D.~Trevor.
\newblock {Fully Convolutional Networks for Semantic Segmentation}.
\newblock In {\em The IEEE Conference on Computer Vision and Pattern
  Recognition (CVPR)}, 2015.

\bibitem{alexnet}
A.~Krizhevsky, I.~Sutskever, and G.~E. Hinton.
\newblock {ImageNet Classification with Deep Convolutional Neural Networks}.
\newblock In {\em Advances in Neural Information Processing Systems (NIPS)},
  pages 1097--1105. Curran Associates, Inc., 2012.

\bibitem{xnor_net}
R.~Mohammad, O.~Vicente, R.~Joseph, and F.~Ali.
\newblock {XNOR-Net: ImageNet Classification Using Binary Convolutional Neural
  Networks}.
\newblock {\em European Conference on Computer Vision (ECCV)}, 2016.

\bibitem{faster_rcnn}
S.~Ren, K.~He, R.~Girshick, and J.~Sun.
\newblock {Faster R-CNN: Towards Real-Time Object Detection with Region
  Proposal Networks}.
\newblock In {\em Neural Information Processing Systems (NIPS)}, 2015.

\bibitem{ilsvrc}
O.~Russakovsky, J.~Deng, H.~Su, J.~Krause, S.~Satheesh, S.~Ma, Z.~Huang,
  A.~Karpathy, A.~Khosla, M.~Bernstein, A.~C. Berg, and L.~Fei-Fei.
\newblock {ImageNet Large Scale Visual Recognition Challenge}.
\newblock {\em International Journal of Computer Vision (IJCV)},
  115(3):211--252, 2015.

\bibitem{greedy_vector_decomposition}
P.~H.~T. Sam~Hare, Amir~Saffari.
\newblock {Efficient Online Structured Output Learning for Keypoint-Based
  Object Tracking}.
\newblock In {\em the Proceedings IEEE Conference of Computer Vision and
  Pattern Recognition (CVPR)}, 2012.

\bibitem{vggnet}
K.~Simonyan and A.~Zisserman.
\newblock {Very Deep Convolutional Networks for Large-Scale Image Recognition}.
\newblock In {\em Proceedings of the International Conference on Learning
  Representations (ICLR)}, 2015.

\bibitem{googlenet}
C.~Szegedy, W.~Liu, Y.~Jia, P.~Sermanet, S.~Reed, D.~Anguelov, D.~Erhan,
  V.~Vanhoucke, and A.~Rabinovich.
\newblock Going deeper with convolutions.
\newblock In {\em The IEEE Conference on Computer Vision and Pattern
  Recognition (CVPR)}, 2015.

\bibitem{segnet}
B.~Vijay, K.~Alex, and C.~Roberto.
\newblock {SegNet: A Deep Convolutional Encoder-Decoder Architecture for Image
  Segmentation}.
\newblock {\em arXiv preprint arXiv:1511.00561}, 2015.

\bibitem{exhaustive_vector_decomposition}
Y.~Yamauchi, A.~Mitsuru, S.~Ikuro, Y.~Yoshida, and H.~Fujiyoshi.
\newblock {Distance Computation Between Binary Code and Real Vector for
  Efficient Keypoint Matching}.
\newblock {\em Information Processing Society of Japan Transactions on Computer
  Vision and Applications (CVA)}, 5:124--128, 2013.

\end{thebibliography}
}

\end{document}